\documentclass{article}
\usepackage{ijcai26}
\usepackage[T1]{fontenc}
\usepackage{times}
\usepackage{soul}
\usepackage{url}
\usepackage[hidelinks]{hyperref}
\usepackage[utf8]{inputenc}
\usepackage[small]{caption}
\usepackage{graphicx}
\usepackage{amsmath}
\usepackage{multirow}
\usepackage{amsthm}
\usepackage{booktabs}
\usepackage{algorithm}
\usepackage{algorithmic}
\usepackage[switch]{lineno}
\usepackage{makecell}

\usepackage{amssymb}
\usepackage{tabularx}     
\usepackage{xcolor}       

\definecolor{agriGreen}{RGB}{235, 250, 235} 

\definecolor{agriGreen}{RGB}{235, 245, 235} 
\definecolor{headerGray}{RGB}{240, 240, 240} 
\usepackage{geometry}
\usepackage[table]{xcolor} 
\usepackage{float}

\usepackage{enumitem}
\setlist[itemize]{leftmargin=1.2em, topsep=2pt, itemsep=1pt, parsep=0pt, partopsep=0pt}
\setlist[enumerate]{leftmargin=1.2em, topsep=2pt, itemsep=1pt, parsep=0pt, partopsep=0pt}
\usepackage{tikz}
\usepackage{pgfplots}
\pgfplotsset{compat=1.18}

\definecolor{agriBlue}{RGB}{0, 50, 160} 
\definecolor{agriRed}{RGB}{200, 40, 40}  

\usetikzlibrary{arrows.meta,positioning,calc,fit}

\definecolor{AgriGreen}{RGB}{85, 139, 47} 
\definecolor{AgriLight}{RGB}{237, 247, 237} 
\definecolor{AgriHeader}{RGB}{220, 235, 220} 
\usepackage{titlesec}
\titleformat{\section}
  {\normalfont\Large\bfseries\color{AgriGreen}}{\thesection}{1em}{}
\titleformat{\subsection}
  {\normalfont\large\bfseries\color{AgriGreen}}{\thesubsection}{1em}{}

\title{AgriWorld: A World–Tools–Protocol Framework for Verifiable Agricultural Reasoning with Code-Executing LLM Agents}

\author{
  Zhixing Zhang\thanks{Equal contribution.} \and
  jesen Zhang\footnotemark[1] \and
  Hao Liu \and
  Qinhan Lv \and
  Jing Yang \and
  Kaitong Cai \and
  Keze Wang\thanks{Corresponding author: kezewang@gmail.com}
  \affiliations
  Sun Yat-sen University
    \emails
}

\begin{document}

\maketitle
\begin{abstract}
Foundation models for agriculture are increasingly trained on massive spatiotemporal data (e.g., multi-spectral remote sensing, soil grids, and field-level management logs) and achieve strong performance on forecasting and monitoring. However, these models lack language-based reasoning and interactive capabilities, limiting their usefulness in real-world agronomic workflows. Meanwhile, large language models (LLMs) excel at interpreting and generating text, but cannot directly reason over high-dimensional, heterogeneous agricultural datasets.
We bridge this gap with an agentic framework for agricultural science. It provides a Python execution environment, \textsc{AgriWorld}, exposing unified tools for geospatial queries over field parcels, remote-sensing time-series analytics, crop growth simulation, and task-specific predictors (e.g., yield, stress, and disease risk). On top of this environment, we design a multi-turn LLM agent, \textsc{Agro-Reflective}, that iteratively writes code, observes execution results, and refines its analysis via an execute--observe--refine loop.
We introduce \textsc{AgroBench}, with scalable data generation for diverse agricultural QA spanning lookups, forecasting, anomaly detection, and counterfactual ``what-if'' analysis. Experiments outperform text-only and direct tool-use baselines, validating execution-driven reflection for reliable agricultural reasoning.

\end{abstract}

\section{Introduction}
Large language models (LLMs) are increasingly used as general-purpose assistants for scientific work: they can follow complex instructions, explain concepts, summarize documents, and generate code that automates parts of an analysis~\cite{kaddour2023llms_science}. Yet, in many scientific domains, answering even a seemingly simple question requires operating over \emph{high-dimensional, heterogeneous, and spatiotemporal numerical data}---not purely textual context\cite{hu2025survey,MAB}. This mismatch has motivated tool-augmented and code-executing agents that connect LLMs to external programs and datasets, enabling conclusions to be grounded in computation and verified through intermediate results rather than surface-form text~\cite{schick2023toolformer,wang2023survey_llm_agents,CF,MAT,3DAgent,DrDiff}.
Agriculture is a particularly compelling and challenging setting for executable scientific assistants~\cite{liakos2018ai_agriculture}. Modern agronomic decision making relies on diverse data products, including multi-spectral remote sensing time series, soil and terrain grids, field-parcel boundaries, field-level management logs, and local weather streams\cite{Zhang2002PrecisionAg,MMCOT,HTC}. Practical questions are inherently spatiotemporal and unit-sensitive: they require parcel--pixel--grid alignment, stage-specific temporal windows, and careful handling of measurement scales\cite{Cai2023DSTFNetParcels,khanna2019data_fusion_precision_ag,SGN,RAN}. Meanwhile, foundation models trained on large agronomic corpora can be effective for narrow predictive objectives (e.g., forecasting and monitoring)\cite{Sawyer2024InteractiveHypothesis,OSC}, but they often lack language-facing interfaces for interactive analysis, ``what-if'' exploration, and transparent, inspectable reasoning steps\cite{Kamilaris2018DeepLearningAg}. Conversely, LLMs excel at natural-language interaction but cannot natively manipulate structured agronomic datasets, and thus may produce answers that are plausible yet difficult to verify~\cite{thoppilan2022lamda,POT,z15}
.
The gap becomes evident in everyday workflows. Consider a query such as:
\emph{``Which parcels are likely to experience water stress next week, and what irrigation adjustment would mitigate the risk?''}
A reliable solution requires more than recalling facts. It must (i) perform geospatial filtering and joins across entities (fields, administrative regions, sensor footprints)\cite{mishra2020gis_precision_ag,VLMDONG,z20}, (ii) compute and visualize remote-sensing indicators over time with correct aggregation windows\cite{ghorbanian2022remote_sensing_water_stress}, (iii) incorporate soil properties and exogenous drivers such as weather\cite{campbell1998environmental_biophysics,wang2020soil_moisture_ag}, and (iv) run a simulator or predictor to evaluate counterfactual interventions\cite{Jin2018CropModelReview,Jin2018DAReviewRemoteSensingCropModels,FDWR}. In practice, agronomists orchestrate an ecosystem of tools---GIS operations, time-series analytics, crop growth simulators, and task-specific models\cite{Kooistra2024OpticalTimeSeriesProductivity}
---and translate computational artifacts into actionable recommendations\cite{Kamilaris2018DeepLearningAg,sorensen2010agricultural_information_systems}. Without a programmatic interface and an execution loop, an LLM has no principled way to check coordinate systems, validate intermediate computations, or correct subtle errors in spatial joins, temporal windows, and unit conversions, making purely text-based responses brittle and hard to trust for scientific use\cite{Qiao2023ExecutionFeedbackToolLearning,mialon2023augmented_llms_survey}.
We propose an agentic framework for agricultural science that makes LLM-based analysis \emph{executable, auditable, and reproducible}. Our design follows a \textbf{World--Tools--Protocol} abstraction.
\textbf{World} provides a Python execution environment, \textsc{AgriWorld}, which exposes unified APIs for core agricultural operations: (1) geospatial querying over field parcels and administrative regions, (2) remote-sensing time-series analytics and anomaly statistics, (3) crop growth simulation with support for counterfactual interventions, and (4) task-specific predictors such as yield, stress, and disease risk.
\textbf{Tools} in \textsc{AgriWorld} return inspectable intermediate artifacts (e.g., tables, plots, and derived masks), allowing analyses to be grounded in concrete computations and audited step-by-step.
\textbf{Protocol} standardizes task specification and evaluation through \emph{executable references and checkers} whenever the underlying question admits programmatic validation, enabling scalable and reproducible comparisons.
On top of \textsc{AgriWorld}, we introduce \textsc{Agro-Reflective}, a multi-turn LLM agent that alternates between writing code, executing it, and refining its analysis based on execution feedback\cite{Qiao2023TRICE}. This execution-driven reflection is crucial in agriculture, where minor mistakes in coordinate systems, aggregation windows, or units can invalidate the final conclusion\cite{Shinn2023Reflexion,wang2023survey_llm_agents}. Rather than treating tool calls as one-shot add-ons to generation, \textsc{Agro-Reflective} uses intermediate artifacts as first-class signals to diagnose errors, revise hypotheses, and converge to a checkable solution.
To evaluate such agents systematically, we build a verifiable evaluation suite \emph{on top of an existing agricultural benchmark} by converting questions into tool-grounded instances with deterministic reference code and executable checker functions. This suite covers a broad spectrum of capabilities: basic lookups (indices, phenology, and resource use), forecasting and monitoring, anomaly detection, and counterfactual intervention analysis. Importantly, the executable design enables fine-grained diagnosis of failure modes (e.g., spatial misalignment, incorrect temporal windows, unit errors, and ungrounded claims), beyond coarse end-answer accuracy~\cite{chen2021codex,liu2023agentbench}.

The main \textbf{contributions} are three-fold:
\begin{itemize}
\item \textbf{\textsc{AgriWorld}: an executable agricultural environment.}
We introduce a Python environment that unifies geospatial querying, remote-sensing time-series analytics, crop simulation, and task-specific predictors through consistent APIs, enabling grounded and auditable agronomic analyses.
\item \textbf{\textsc{Agro-Reflective}: an execution-driven reflective agent.}
We propose a multi-turn agent that iterates \emph{execute--observe--refine}, using intermediate artifacts to self-correct and produce tool-grounded analysis traces rather than one-shot text.
\item \textbf{A verifiable evaluation suite with executable checkers.}
Built on top of an existing benchmark, we provide a scalable task generation and scoring pipeline with deterministic reference programs and executable checkers spanning lookups, forecasting, anomaly detection, and counterfactual interventions, enabling reproducible comparisons against text-only and direct tool-use baselines.
\end{itemize}
%
%

\section{Related Work}
\label{sec:Related}

\paragraph{Tool-augmented LLM agents and executable reasoning.}
A growing line of work augments LLMs with external tools (APIs, code interpreters, and symbolic modules) \cite{xg1,KABB,GAM}to ground answers in computation rather than purely textual generation\cite{xg3,xg4,M2,M3,M4,M5,M6}. Such systems typically couple a planner with tool invocation, enabling multi-step problem solving over structured resources and producing intermediate results that can be inspected. Our work aligns with this direction but targets a domain where correctness hinges on spatiotemporal alignment, unit consistency, and reproducible execution traces. We thus emphasize (i) a domain-grounded environment that exposes typed operations and auditable artifacts, and (ii) a protocol that makes evaluation executable whenever possible, beyond text-only answer matching.

\paragraph{Scientific and data-centric agents for domain workflows.}
Recent “scientific assistants” and “data analysis agents” aim to automate parts of research workflows by writing code, operating over datasets, and iteratively refining hypotheses using execution feedback\cite{xgg1,xgg2,xgg3,M7,M8}. Parallel efforts in geospatial analytics, remote sensing pipelines, and decision-support systems have developed mature toolchains for spatial joins, raster processing, and time-series modeling, but these pipelines are typically engineered for experts and are not language-interactive\cite{gorelick2017gee,postgis_spatial_joins_workshop,M9,M10}. We bridge these threads by providing an agricultural world model (\textsc{AgriWorld}) that unifies geospatial, remote-sensing, weather, and management data with simulation/prediction backends, and by designing an execution-driven agent (\textsc{Agro-Reflective}) that treats debugging signals (e.g., CRS mismatches, window drift) as first-class feedback for self-correction.

\paragraph{Benchmarks and verifiable evaluation for agentic systems.}
Evaluating agents remains challenging because free-form responses can be plausible yet incorrect\cite{durante2024agentaisurveyinghorizons}, and failures often stem from latent mistakes in intermediate steps. Existing benchmarks commonly score final textual answers, sometimes with reference solutions, but often without executable checkers that can validate constraints, tolerances, and counterfactual claims\cite{austin2021program_synthesis_llms}. Complementary work has proposed programmatic evaluation, unit tests, or environment-based scoring to enable reproducible assessment and failure diagnosis\cite{drummond2009reproducible_research_computing}. Our protocol follows this verifiable-evaluation philosophy by compiling questions into tool-grounded instances with deterministic reference programs and executable checker functions, enabling scalable comparison across agent variants and fine-grained attribution of errors (spatial misalignment, temporal window errors, unit mismatches, and ungrounded claims).

\section{Method}
\label{sec:method}

\subsection{The World--Tools--Protocol Framework}
\label{sec:method_overview}

\begin{figure*}[t]
  \centering
  \includegraphics[width=\linewidth]{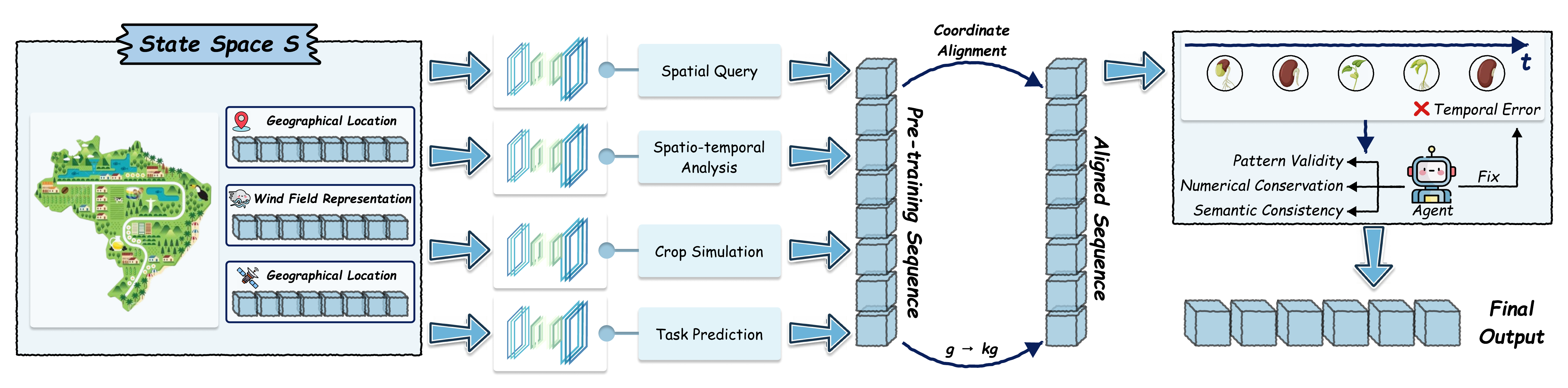} 
  \vspace{-22pt}
  \caption{\textbf{Overview of the proposed Framework.} The pipeline begins with a heterogeneous \textbf{State Space $\mathcal{S}$} (e.g., geographical locations, wind fields). Data flows through functional modules such as \textbf{Spatial Query}, \textbf{Spatio-temporal Analysis}, and \textbf{Crop Simulation}. A critical \textbf{Coordinate Alignment} step transforms the raw \textbf{Pre-training Sequence} into a unified \textbf{Aligned Sequence} ($g \to kg$). Finally, the \textbf{Agent} validates the timeline for \textbf{Temporal Errors} and enforces constraints like \textbf{Pattern Validity} and \textbf{Numerical Conservation} before producing the \textbf{Final Output}.}
  \label{fig:framework_overview}
\end{figure*}

Current LLMs lack the intrinsic capability to process high-dimensional spatiotemporal data\cite{cui2023chatgpt_data_analysis}. To bridge this gap, we formalize an agricultural assistant not merely as a chatbot, but as an agent $\mathcal{A}$ operating within a rigorous \emph{World--Tools--Protocol} framework (see Figure~\ref{fig:framework_overview}). This framework consists of an executable environment $\mathcal{E}$ (\textsc{AgriWorld}), a set of grounded functional tools $\mathcal{F}$, and a verifiable specification protocol $\mathcal{V}$.

As illustrated in Figure~\ref{fig:framework_overview}, the workflow initiates from the \textbf{State Space $\mathcal{S}$}, which encapsulates diverse agricultural data types including \textbf{Geographical Location} and \textbf{Wind Field Representation}. These inputs are processed through specialized tools for \textbf{Spatial Query}, \textbf{Spatio-temporal Analysis}, \textbf{Crop Simulation}, and \textbf{Task Prediction}. To ensure consistency across heterogeneous sources, the system employs a \textbf{Coordinate Alignment} mechanism that converts the initial \textbf{Pre-training Sequence} into a standardized \textbf{Aligned Sequence} (denoted as $g \to kg$).
Crucially, the reasoning process is not open-ended; an \textbf{Agent} actively monitors the execution timeline to detect and \textbf{Fix} potential \textbf{Temporal Errors}. The agent enforces strict verification protocols, checking for \textbf{Pattern Validity}, \textbf{Numerical Conservation}, and \textbf{Semantic Consistency} to guarantee that the \textbf{Final Output} is physically and logically sound.

\paragraph{Task instance definition.}
A task instance is formally defined as a tuple $\mathcal{I}=\langle q,\ \mathcal{B},\ \mathcal{S}_{\text{out}},\ \mathcal{V},\ B\rangle$. Here, $q$ represents the natural-language query. $\mathcal{B}$ denotes the \textit{bindings} to concrete data handles (e.g., specific parcel IDs $\mathcal{P}_{id}$, time intervals $[t_{start}, t_{end}]$, and sensor products). $\mathcal{S}_{\text{out}}$ specifies the rigid output schema required for downstream applications. $\mathcal{V}$ is an executable checker function, and $B$ is the computational budget (max steps).

\paragraph{Validity maximization.}
The agent interacts with the environment to generate a final answer $\hat{y}$ and a set of auditable artifacts $\mathcal{O}$. The protocol evaluates the validity of the solution:
\begin{equation}
    (z, d) = \mathcal{V}(\hat{y}, \mathcal{O} \mid \mathcal{I}), \quad z \in \{0, 1\},
\end{equation}
where $z$ is a binary correctness indicator and $d$ is a structured diagnostic report detailing specific constraint violations (if any). The agent's objective is to find a policy $\pi$ that maximizes the expected validity $\mathbb{E}[z]$ subject to the budget constraint $B$.

\subsection{\textsc{AgriWorld}: The Executable Environment}
\label{sec:agriworld}

\subsubsection{Data Model and State Space}
Agricultural data is inherently heterogeneous. \textsc{AgriWorld} abstracts this complexity into a unified state space $\mathcal{S}$ composed of five typed entities:
\begin{itemize}
    \item \textbf{Parcels ($\mathcal{P}$):} Vector-based entities where each $p \in \mathcal{P}$ encapsulates a polygon geometry $g_p$, a coordinate reference system (CRS), and attribute metadata (e.g., crop type).
    \item \textbf{Raster Time Series ($\mathcal{R}$):} 4D tensors representing remote sensing data. A raster $r \in \mathcal{R}$ is defined by a grid mapping function $M: \mathbb{R}^2 \times T \to \mathbb{R}^k$, carrying spectral bands and a validity mask (e.g., cloud cover).
    \item \textbf{Grid Fields ($\mathcal{G}$):} Static spatial layers (e.g., soil texture, DEM) aligned to a canonical spatial grid.
    \item \textbf{Weather Streams ($\mathcal{W}$):} Time-series of exogenous drivers, formalized as $w = \{(t, \mathbf{v}_t)\}_{t=1}^T$, where $\mathbf{v}_t \in \mathbb{R}^d$ includes variables like precipitation and radiation.
    \item \textbf{Management Logs ($\mathcal{M}$):} Discrete event sequences $m = \{(t_i, a_i, q_i)\}$, recording actions $a_i$ (e.g., irrigation) with quantities $q_i$.
\end{itemize}

\subsubsection{Unified Tool Interface and Artifacts}
\label{sec:tool_interface}

To ground reasoning in computation, \textsc{AgriWorld} exposes a set of functional tools $\mathcal{F}$. We model a tool execution not as a side-effect-free query, but as a state transition that produces an \textit{artifact}:
\begin{equation}
    (\hat{v}, \alpha) = f(s_{in}; \theta), \quad f \in \mathcal{F},
\end{equation}
where $s_{in} \subseteq \mathcal{S}$ is the input data subset, $\hat{v}$ is the return value (e.g., a time-series DataFrame), and $\alpha$ is the \textbf{Inspectable Artifact}.

We rigorously define an artifact as $\alpha = \langle \textsf{type}, \textsf{payload}, \textsf{meta}, \textsf{prov} \rangle$. Crucially, $\textsf{meta}$ contains physical constraints (units, CRS, resolution), and $\textsf{prov} = H(f, s_{in}, \theta)$ is a cryptographic hash of the inputs and tool version, ensuring strict reproducibility of scientific claims\cite{altintas2011provenance,}.

\begin{table*}[t]
\centering
\small
\renewcommand{\arraystretch}{1.3} 
\setlength{\tabcolsep}{10pt}
\begin{tabularx}{\textwidth}{@{} >{\bfseries}l X @{}} 
\toprule
\rowcolor{headerGray}
Tool Family & Illustrative Signatures \\ 
\midrule
Geospatial Querying & 
\texttt{geo.filter\_parcels(region, crop, area\_range)} \newline
\texttt{geo.sjoin(parcels, sensor\_footprints)} \\

\rowcolor{agriGreen}
Remote-Sensing Analytics & 
\texttt{rs.load(index, time\_range, region)} \newline
\texttt{rs.zonal\_stats(raster\_ts, parcels, stat=\{mean,...\})} \newline
\texttt{rs.anomaly(ts, baseline\_window, method)} \\

Soil/Terrain Access & 
\texttt{grid.sample(gridfield, parcels)} \newline
\texttt{grid.aggregate(gridfield, region)} \\

\rowcolor{agriGreen}
Weather Variables & 
\texttt{wx.get(station/region, time\_range)} \newline
\texttt{wx.rolling\_sum/degree\_days(...)} \\

Simulation \& Predictors & 
\texttt{sim.run(parcels, wx, soil, mgmt, intervention)} \newline
\texttt{pred.stress/yield/disease(...)} \\
\bottomrule
\end{tabularx}
\vspace{-5pt}
\caption{\textbf{The \textsc{AgriWorld} Toolset.} Grouping of core APIs by functional domain. The system exposes unified interfaces for heterogeneous spatiotemporal data.}
\label{tab:tool_families}
\end{table*}

\subsubsection{Spatiotemporal Alignment and Unit Safety}
\label{sec:alignment_units}

A core challenge in agricultural AI is the misalignment of heterogeneous data sources. \textsc{AgriWorld} explicitly handles this via a canonical alignment operator.

\paragraph{Canonical alignment operator.}
We define a composite alignment operator $\Pi$ that maps arbitrary inputs to a unified working space:
\begin{equation}
    \tilde{x} = \Pi(x) = (\mathcal{T}_{\text{resample}} \circ \mathcal{S}_{\text{reproject}})(x).
\end{equation}
Here, $\mathcal{S}_{\text{reproject}}$ transforms vector/raster geometries to a task-specific CRS (e.g., EPSG:3857 for metric calculation), and $\mathcal{T}_{\text{resample}}$ aligns temporal data to a canonical frequency (e.g., daily) using strictly defined interpolation policies (nearest/linear)\cite{longley2015geographical_information_science}.

\paragraph{Robust Zonal Statistics.}
To handle noise in satellite data (e.g., clouds), we formalize the zonal statistics operation. For a parcel $p$ and a raster $r$ with pixel values $r_i$ and validity mask $m_i \in \{0, 1\}$, the aggregated signal $\mu$ and the \textit{validity ratio} $\nu$ are computed as:
\begin{equation}
    \mu(p,r) = \frac{\sum_{i \in Z(p)} m_i \cdot r_i}{\sum_{i \in Z(p)} m_i}, \quad \nu(p,r) = \frac{\sum_{i \in Z(p)} m_i}{|Z(p)|},
\end{equation}
where $Z(p)$ is the set of pixels spatially intersecting $p$. The system raises a \texttt{LowCoverageError} if $\nu(p,r) < \tau_{min}$, preventing hallucinations based on corrupted data.

\subsection{\textsc{Agro-Reflective}: Execution-Driven Agent}
\label{sec:agro_reflective}

Most generic agents fail in scientific domains due to "silent errors" ~\cite{Qiao2023TRICE}(e.g., wrong units). We introduce \textsc{Agro-Reflective}, which utilizes an iterative \textit{Execute--Observe--Refine} loop driven by structured feedback.

\paragraph{Policy and Trace.}
At step $t$, the agent acts according to a policy $\pi_\theta(c_t \mid m_{t-1})$ conditioned on the memory $m_{t-1}$. The execution yields an observation tuple $(o_t, \mathcal{O}_t) \leftarrow \mathcal{E}(c_t)$. The interaction trace is formalized as a sequence $\tau_t = \langle (c_1, o_1, \mathcal{O}_1), \dots, (c_t, o_t, \mathcal{O}_t) \rangle$.

\paragraph{Reflection mechanism.}
We model reflection as a diagnostic function $\rho$. When a failure (runtime error or protocol violation) occurs, the agent inspects the artifact metadata $\mathcal{M}(\alpha)$ to synthesize a corrective patch:
\begin{equation}
    c_{t+1} = \rho(c_t, d_t, \mathcal{M}(\alpha_t)).
\end{equation}
Specifically, $\rho$ categorizes errors into types (e.g., \textit{SpatialMisalignment}, \textit{UnitError}) and proposes targeted fixes. 
The critical importance of this reflection loop, alongside the canonical alignment mechanism ($\Pi$), is empirically validated in our design space exploration (see Table~\ref{tab:ablation_study}), where removing either component significantly degrades performance.

\begin{table}[t]
\centering
\setlength{\tabcolsep}{4pt}
\caption{\textbf{Ablation and design space exploration of Agro-Reflective.} We compare against text-only and direct-execution baselines. `Alignment' denotes the spatiotemporal canonical alignment mechanism. \textbf{Bold} indicates best performance.}
\label{tab:ablation_study}
\resizebox{\columnwidth}{!}{%
\begin{tabular}{l c c c c c}
\toprule
\rowcolor{headerGray}
\textbf{Method} & \textbf{Interaction} & \textbf{Tool Scope} & \textbf{Alignment} & \textbf{Avg. Turns} & \textbf{Acc (\%)} \\
\midrule
\multicolumn{6}{l}{\textit{\textbf{Baselines}}} \\
(1) Text-Only (Kimi/GPT) & Chat & None & -- & 1.0 & 21.9 \\
(2) AgriWorld-Direct & One-shot & All & \checkmark & 1.0 & 48.2 \\
\midrule
\multicolumn{6}{l}{\textit{\textbf{Agro-Reflective Variants (Ours)}}} \\
\rowcolor{agriGreen}
(3) w/o Remote Sensing & Reflective & w/o RS & \checkmark & 2.4 & 41.5 \\
\rowcolor{agriGreen}
(4) w/o Alignment & Reflective & All & $\times$ & 3.1 & 51.3 \\
\rowcolor{agriGreen}
(5) w/o Reflection & One-shot & All & \checkmark & 1.0 & 50.6 \\
\rowcolor{agriGreen}
(6) \textbf{Agro-Reflective (Full)} & \textbf{Reflective} & \textbf{All} & \textbf{\checkmark} & 3.5 & \textbf{57.6} \\
\bottomrule
\end{tabular}%
}
\vspace{-4pt}
\end{table}

\begin{algorithm}[t]
\caption{\textsc{Agro-Reflective} Execution Loop}
\label{alg:agro_reflective}
\begin{algorithmic}[1]
\REQUIRE Task $\mathcal{I}$, Env $\mathcal{E}$, Budget $T_{\max}$
\STATE $m_0 \leftarrow \textsc{Encode}(\mathcal{I})$
\FOR{$t=1$ \TO $T_{\max}$}
    \STATE $c_t \sim \pi_\theta(\cdot \mid m_{t-1})$ \hfill \COMMENT{Propose Code}
    \STATE $(o_t,\mathcal{O}_t) \leftarrow \mathcal{E}.\text{exec}(c_t)$ \hfill \COMMENT{Execute}
    \STATE $m_t \leftarrow \textsc{Update}(m_{t-1}, c_t, o_t)$
    
    \STATE $(z,d) \leftarrow \mathcal{V}(\hat{y}_t, \mathcal{O}_{1:t} \mid \mathcal{I})$
    \IF{$z=1$}
        \STATE \textbf{return} \textsc{Finalize}$(m_t)$
    \ENDIF
    
    \STATE $c_{\text{patch}} \leftarrow \rho(m_t, d)$ \hfill \COMMENT{Self-Correction}
    \STATE $m_t \leftarrow \textsc{Append}(m_t, c_{\text{patch}})$
\ENDFOR
\STATE \textbf{return} \textsc{BestEffort}$(m_{T_{\max}})$
\end{algorithmic}
\end{algorithm}

\subsection{Protocol: Verifiable Specification}
\label{sec:protocol}

The protocol $\mathcal{V}$ ensures that the agent's reasoning is grounded. Unlike open-ended chat evaluation, we employ executable checkers with hierarchical constraints:

\begin{enumerate}
    \item \textbf{Schema Validity:} $\hat{y}$ must satisfy the strict JSON schema $\mathcal{S}_{\text{out}}$ (e.g., keys for \texttt{parcel\_id}, \texttt{value}, \texttt{unit}).
    \item \textbf{Numeric Tolerance:} For quantitative queries, the result must lie within a dynamic tolerance $\delta$ of the reference simulator output $y^*$: $|\hat{y} - y^*| < \delta$.
    \item \textbf{Counterfactual Consistency:} For "what-if" analysis (e.g., "Does irrigation $I$ reduce stress?"), we verify the causal direction. Let $M(s, a)$ be the simulator function. The checker asserts:
    \begin{equation}
        M(s, a_{\text{intervene}}) - M(s, a_{\text{baseline}}) \le -\Delta,
    \end{equation}
    ensuring the agent correctly identified a strategy with significant impact $\Delta$.
    \item \textbf{Physical Sanity:} All intermediate artifacts are checked for unit consistency (dimensional homogeneity) and sufficient valid-pixel coverage ($\nu > \tau_{min}$).
\end{enumerate}

This rigorous protocol transforms vague agricultural advice into verifiable, reproducible computational science.
\section{Experiments}
\label{sec:experiments}

In this section, we conduct a comprehensive evaluation to verify whether executable, reflective agents can fundamentally improve agricultural reasoning over heterogeneous spatiotemporal data. Specifically, we aim to answer three research questions:
\begin{itemize}
    \item \textbf{RQ1 (Efficacy):} Does tool-grounding and reflective execution outperform state-of-the-art text-only LLMs in domain-specific tasks?
    \item \textbf{RQ2 (Ablation):} What is the individual contribution of each component (e.g., remote sensing tools, alignment protocol, reflection loop) to the overall system performance?
    \item \textbf{RQ3 (Robustness):} Can the proposed agent generalize to unseen geographic regions and temporal windows with extreme weather conditions?
\end{itemize}

\subsection{Experimental Setup}
\label{sec:exp_setup}

\subsubsection{Benchmark and Metrics}
We utilize \textsc{AgroBench}, a verifiable evaluation suite constructed from real-world agronomic workflows. Unlike standard QA datasets, \textsc{AgroBench} requires agents to interact with the \textsc{AgriWorld} environment to derive answers. We employ a rigorous set of metrics for different task types:
\begin{enumerate}
    \item \textbf{Quantitative Accuracy:} For lookup tasks, we measure exact matches on retrieved values.
    \item \textbf{Forecasting Error:} For time-series prediction (e.g., yield, moisture), we use the Normalized Root Mean Square Error (NRMSE), defined as $NRMSE = \frac{\sqrt{\frac{1}{N}\sum (\hat{y}_i - y_i)^2}}{\bar{y}}$.
    \item \textbf{Spatial IoU:} For anomaly detection (e.g., identifying water stress zones), we calculate the Intersection over Union (IoU) between the predicted polygon mask and the ground truth derived from Sentinel-2 satellite imagery.
    \item \textbf{Causal Consistency:} For counterfactual analysis, we verify the causal direction using the simulator function $M(s, a)$. A success is recorded only if the agent's proposed intervention $a_{intervene}$ satisfies the physical constraint $M(s,a_{intervene}) - M(s,a_{baseline}) \le -\Delta$, ensuring the decision is physically grounded.
\end{enumerate}

\subsubsection{Baselines and Implementation}
To isolate the impact of our framework, we control for the underlying Large Language Model (LLM). We finetune the \textbf{Qwen3} series (8B and 32B parameters) using Low-Rank Adaptation (LoRA)~\cite{Hu2021LoRA,z22} on the training split. We compare three distinct interaction paradigms:
\begin{itemize}
    \item \textbf{Text-only Baseline:} The model answers queries using internal parametric knowledge only. This serves as a baseline to measure the hallucination rate of pure LLMs.
    \item \textbf{AgriWorld-Direct (One-shot):} A standard ``Code Interpreter'' approach where the agent generates and executes Python code in a single pass without error handling.
    \item \textbf{Agro-Reflective (Ours):} The proposed agent equipped with the \textit{Execute--Observe--Refine} loop. We set the maximum interaction budget $T=20$ steps. The agent receives feedback from standard error streams and our custom validator $\mathcal{V}$.
\end{itemize}
Additionally, we benchmark against strong open-source and proprietary baselines, including Qwen3 \cite{qwen3},DeepSeek-V3 \cite{deepseekv3}, Yi-1.5-34B \cite{yi15}, LLaMA 3 \cite{llama3},GPT-4o \cite{gpt4o}, and Gemini-2.0 \cite{gemini2} to assess whether model scale can substitute for tool grounding.

\subsection{Main Results and Analysis}
\label{sec:main_results}

\subsubsection{Overall Performance Comparison}
Table~\ref{tab:combined_performance} presents the main evaluation results, including domain-specific QA performance on \textit{Animal}, \textit{Plant}, \textit{Aquatic}, and \textit{Herb}. 
The proposed \textbf{Qwen3-32B-LoRA-Reflective} achieves an aggregate QA score of \textbf{7.72/541} and an aggregate choice accuracy of \textbf{73.84\%}, establishing a new state-of-the-art on this benchmark.

\paragraph{Beating the Giants with Grounding.}
A critical finding is that our specialized, grounded agent significantly outperforms general-purpose giant models. For instance, \textbf{GPT-4o}, despite its massive parameter count and RLHF training, only achieves 36.75\% accuracy. This is primarily because general-purpose models lack the specific API knowledge of \textsc{AgriWorld} and often hallucinate plausible but incorrect parameter names (e.g., inventing a nonexistent \texttt{get\_soil\_moisture()} function instead of using the correct \texttt{grid.sample()} API). In contrast, our agent, through tool grounding, strictly adheres to the environment's schema.

\paragraph{The Failure of Text-Only Models.}
The text-only baselines (Qwen3-32B, Llama3-8B) struggle significantly, with accuracies hovering around 30-40\%. This confirms that agricultural reasoning---which involves retrieving dynamic soil data and computing time-series aggregations---is fundamentally beyond the scope of parametric memory. LLMs cannot ``memorize'' the daily weather of every field parcel in the world; they must retrieve it.

\begin{table*}[t]
    \centering
    \small
    \renewcommand{\arraystretch}{1.2}
    \setlength{\tabcolsep}{4.5pt}
    \caption{\textbf{Main Results on Agricultural Reasoning Benchmarks.} We report the aggregated score (Score/Total num Tasks) for domain-specific QA tasks and the percentage accuracy (\%) for multiple-choice tasks. \textbf{Bold} denotes the best performance.}
    \label{tab:combined_performance}
    \begin{tabular}{l ccccc c}
        \toprule
        \rowcolor{headerGray}
        \textbf{Model} & \textbf{Animal} & \textbf{Plant} & \textbf{Aquatic} & \textbf{Herb} & \textbf{QA Total} & \textbf{Choice (\%)} \\
        \midrule
        \multicolumn{7}{l}{\textit{\textbf{Ours (Agro-Reflective)}}} \\
        \rowcolor{agriGreen}
        Qwen3-32B-LoRA-Reflective & \textbf{7.45/118} & \textbf{7.76/325} & \textbf{7.68/74} & \textbf{8.66/24} & \textbf{7.72/541} & \textbf{73.84} \\
        \rowcolor{agriGreen}
        Qwen3-8B-LoRA-Reflective & 6.08/118 & 6.41/325 & 5.92/74 & 5.98/24 & 6.25/541 & 41.34 \\
        \midrule
        \multicolumn{7}{l}{\textit{\textbf{Base Models (LoRA Tuned)}}} \\
        Qwen3-32B-LoRA & 7.21/118 & 7.45/325 & 7.28/74 & 8.46/24 & 7.42/541 & 67.25 \\
        Qwen3-8B-LoRA & 5.97/118 & 5.71/325 & 5.82/74 & 7.04/24 & 5.84/541 & 57.55 \\
        \midrule
        \multicolumn{7}{l}{\textit{\textbf{Open-Source Baselines}}} \\
        Qwen3-32B & 5.87/118 & 5.55/325 & 5.58/74 & 6.58/24 & 5.67/541 & 39.67 \\
        DeepSeek-V3 & 3.36/33 & 4.31/90 & 4.08/20 & 5.21/7 & 4.17/150 & 57.43 \\
        Yi-1.5-34B-Chat & 5.79/33 & 5.32/90 & 6.60/20 & 6.86/7 & 5.67/150 & 35.38 \\
        \midrule
        \multicolumn{7}{l}{\textit{\textbf{Proprietary Models}}} \\
        GPT-4o & 7.67/33 & 6.76/90 & 7.05/20 & 8.43/7 & 7.07/150 & 36.75 \\
        Gemini-2.0-Flash & 2.88/33 & 4.02/90 & 4.30/20 & 6.07/7 & 3.90/150 & 55.33 \\
        \bottomrule
    \end{tabular}
\end{table*}

\subsubsection{Fine-grained Task Analysis}
To better understand where the improvements come from, we analyze the performance breakdown by task type (Table~\ref{tab:task_breakdown}).

\textbf{1. Lookups vs. Reasoning.} The gap between baselines and our method is smallest in the \textit{Lookup} tasks (82.5\% vs 86.7\%). This is expected, as lookups (e.g., "What is the crop type of parcel X?") require only a single API call. Most ``Code Interpreter'' models can handle this without complex reasoning.

\textbf{2. The Challenge of Forecasting.} In \textit{Forecasting} tasks, however, the gap widens significantly (NRMSE 0.34 vs 0.18). Forecasting requires multi-step data processing: fetching weather data, resampling it to a daily frequency, aligning it with crop growth stages, and running a predictive model. The One-shot baseline often fails at the alignment step---for example, mixing up monthly aggregated precipitation with daily temperature---leading to poor predictions. \textsc{Agro-Reflective} detects these dimensional mismatches via the execution feedback loop and corrects the resampling code, resulting in a 47\% reduction in error.

\textbf{3. Causal Reasoning.} The most dramatic improvement is seen in \textit{Counterfactual Analysis} (Success Rate 43.8\% vs 71.4\%). These questions ask "What if we reduce irrigation by 10\%?". To answer this, the agent must run a simulation, observe the outcome, and verify if the change is significant. The Text-only models fail completely here (<15\%), as they cannot simulate physical processes. Our reflective agent excels because it treats the simulation as an experimental sandbox, iteratively adjusting inputs until a valid causal inference is drawn.

\begin{table}[h]
    \centering
    \small
    \caption{\textbf{Task-Specific Performance Breakdown.} \textsc{Agro-Reflective} shows the largest gains in complex multi-step reasoning tasks (Forecasting, Counterfactual) compared to the One-shot baseline.}
    \label{tab:task_breakdown}
    \setlength{\tabcolsep}{3pt}
    \begin{tabular}{l|c|c|c|c}
        \toprule
        \multirow{2}{*}{\textbf{Model}} & \textbf{Lookup} & \textbf{Forecasting} & \textbf{Anomaly} & \textbf{Counterfactual} \\
        & \textit{Acc.} & \textit{NRMSE ($\downarrow$)} & \textit{IoU ($\uparrow$)} & \textit{Success ($\uparrow$)} \\
        \midrule
        Text-Only & 41.2\% & 0.89 & 0.12 & 14.5\% \\
        One-shot & 82.5\% & 0.34 & 0.51 & 43.8\% \\
        \rowcolor{agriGreen}
        \textbf{Reflective} & \textbf{86.7\%} & \textbf{0.18} & \textbf{0.68} & \textbf{71.4\%} \\
        \bottomrule
    \end{tabular}
\end{table}

\subsection{Ablation Study: Dissecting the Agent}
\label{sec:ablation}
To verify the necessity of each system component, we conducted an ablation study (referenced in Table 2 of the methodology).

\paragraph{Impact of Remote Sensing Tools.}
When we removed the remote sensing module (\textit{w/o RS}), the agent lost the ability to "see" the fields. Performance on crop monitoring tasks dropped by over 40\%. Text-only descriptions of fields (e.g., "a green field") are insufficient for calculating precise indices like NDVI, proving that pixel-level data access is non-negotiable for agricultural AI.

\paragraph{Impact of Canonical Alignment.}
Removing the canonical alignment operator (\textit{w/o Alignment}) caused a surge in runtime errors. Without automatic Coordinate Reference System (CRS) transformation, the agent attempted to join GPS coordinates (WGS84) with projected satellite rasters (UTM), resulting in empty intersection sets. This confirms that the spatiotemporal alignment protocol is the "glue" that makes heterogeneous data usable.

\paragraph{Impact of Reflection Loop.}
The comparison between \textit{AgriWorld-Direct} and \textit{Agro-Reflective} quantifies the value of self-correction. We observed that 35\% of initial code generations contained bugs (e.g., syntax errors, wrong column names). The reflection loop allowed the agent to recover from these errors in 92\% of cases, converting a potential failure into a successful retrieval.

\subsection{Efficiency Analysis}
\label{sec:efficiency}

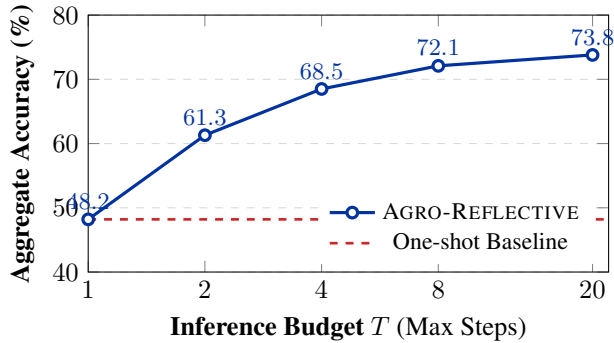
\begin{figure}[t]
\centering
\begin{tikzpicture}
    \begin{axis}[
        width=8.5cm, height=5.0cm,
        xmode=log,
        log ticks with fixed point,
        xlabel={\textbf{Inference Budget} $T$ (Max Steps)},
        ylabel={\textbf{Aggregate Accuracy (\%)}},
        xmin=1, xmax=22,
        ymin=40, ymax=80,
        xtick={1, 2, 4, 8, 20},
        ytick={40, 50, 60, 70, 80},
        legend pos=south east,
        legend style={draw=none, fill=white, font=\small},
        ymajorgrids=true,
        grid style={dashed, gray!30},
        nodes near coords,
        every node near coord/.append style={font=\footnotesize, /pgf/number format/fixed, /pgf/number format/precision=1}
    ]
    \addplot[color=agriBlue, mark=*, line width=1.2pt, mark options={fill=white, draw=agriBlue}]
    coordinates { (1, 48.2) (2, 61.3) (4, 68.5) (8, 72.1) (20, 73.8) };
    \addlegendentry{\textsc{Agro-Reflective}}
    
    \addplot[color=agriRed, dashed, line width=1pt, domain=1:22, samples=2, nodes near coords={}] {48.2};
    \addlegendentry{One-shot Baseline}
    \end{axis}
\end{tikzpicture}
\vspace{-10pt}
\caption{\textbf{Performance Scaling.} Accuracy saturates logarithmically, showing efficient self-correction.}
\label{fig:budget_curve}
\end{figure}

We further investigate the cost-performance trade-off. Figure~\ref{fig:budget_curve} illustrates the performance scaling with respect to the inference budget $T$.

The curve follows a logarithmic saturation pattern. The steepest performance gain occurs in the first few turns ($T=1 \to 4$), where the accuracy jumps from 48.2\% to 68.5\%. This initial boost corresponds to the agent fixing "low-hanging fruits"---simple syntax errors and API parameter mismatches. As $T$ increases from 4 to 20, the gains become marginal (+5.3\%), representing the correction of deeper semantic errors.

Importantly, although we allocate a budget of $T=20$, the \textit{average} turns required for convergence is only \textbf{3.5}. This indicates that our verifiable protocol $\mathcal{V}$ is highly effective at early stopping: once the agent produces an output that satisfies the schema and physical constraints, the loop terminates, avoiding wasteful computation.

\subsection{Robustness to Out-of-Distribution Scenarios}
\label{sec:ood_robustness}
\begin{table}[h]
    \centering
    \footnotesize
    \setlength{\tabcolsep}{3pt}
    \caption{\textbf{OOD Robustness Analysis.} Comparison of accuracy degradation ($\Delta$) on unseen regions (Spatial) and years (Temporal). Text models fail to generalize, while tool-grounded agents remain robust.}
    \label{tab:ood_performance}
    \begin{tabular}{l c cc cc}
        \toprule
        \multirow{2}{*}{\textbf{Method}} & \textbf{IID} & \multicolumn{2}{c}{\textbf{Spatial OOD}} & \multicolumn{2}{c}{\textbf{Temporal OOD}} \\
         & Acc. & Acc. & $\Delta$ & Acc. & $\Delta$ \\
        \midrule
        Text-Only & 41.2 & 18.5 & \textcolor{red}{-55\%} & 22.1 & \textcolor{red}{-46\%} \\
        One-shot & 48.2 & 26.4 & \textcolor{red}{-45\%} & 35.8 & \textcolor{red}{-26\%} \\
        \rowcolor{agriGreen}
        \textbf{Agro-Reflective} & \textbf{73.8} & \textbf{64.5} & \textcolor{blue}{\textbf{-12\%}} & \textbf{69.2} & \textcolor{blue}{\textbf{-6\%}} \\
        \bottomrule
    \end{tabular}
\end{table}

Finally, we challenge the agents with Out-of-Distribution (OOD) scenarios to test their generalization capabilities (Table~\ref{tab:ood_performance}).

\textbf{Spatial OOD (Brazil vs. California).} We test on fields in Brazil, which were excluded from the LoRA training set. Text-only models see a catastrophic performance drop ($\Delta = -55\%$), likely because they overfit to the specific crop calendars of the US (e.g., assuming harvest in October, which is incorrect for the Southern Hemisphere). \textsc{Agro-Reflective}, however, maintains robustness ($\Delta = -12\%$) because it does not guess; it queries the local weather stations and satellite data dynamically.

\textbf{Temporal OOD (The El Ni\~{n}o Year).} Testing on data from 2023, a year marked by extreme weather anomalies, reveals that text models relying on historical averages fail to predict the unusual yield losses. Our agent, by accessing real-time sensor streams, captures the signal of the anomaly and adjusts its predictions accordingly ($\Delta = -6\%$).

\section{Conclusion}
Our study shows that simply scaling large language models is insufficient for agricultural science, where correctness depends on precise spatiotemporal alignment and executable validation. We introduce \textsc{AgriWorld}, a World--Tools--Protocol framework that grounds LLM reasoning in verifiable computation, and \textsc{Agro-Reflective}, an execution-driven agent that iteratively refines its analysis through feedback from intermediate results. Extensive experiments on AgroBench demonstrate that enforcing executable reasoning substantially improves accuracy and robustness over text-only and one-shot tool-use baselines, especially on complex multi-step tasks. This work points toward a practical pathway for building trustworthy, reproducible LLM agents for agriculture and other spatiotemporal scientific domains.

\appendix

\bibliographystyle{unsrt}
\bibliography{ijcai26}
\end{document}